%% file: main_journal.tex
\documentclass[sn-basic]{sn-jnl}

\usepackage{graphicx}%
\usepackage{multirow}%
\usepackage{amsmath,amssymb,amsfonts}%
\usepackage{amsthm}%
\usepackage{mathrsfs}%
\usepackage[title]{appendix}%
\usepackage{xcolor}%
\usepackage{textcomp}%
\usepackage{manyfoot}%
\usepackage{booktabs}%
\usepackage{algorithm}%
\usepackage{algorithmicx}%
\usepackage{algpseudocode}%
\usepackage{listings}%
\usepackage{tikz}
\usepackage{pgfplots}
\usepackage{subcaption}
\pgfplotsset{compat=1.18}
\usepgfplotslibrary{groupplots}
\pgfplotsset{
  every axis/.append style={
    scale only axis,
  }
}

\usepackage{float}
\usepackage{amsmath}

\newcommand{\PhytoNode}{PhytoNode}
\newcommand{\Aust}{Aust}

\definecolor{question}{RGB}{255,165,0}
\newcommand\ebq[1]{\textcolor{question}{#1}}
\newif\ifextended
\extendedfalse     

\newcommand{\ext}[1]{\ifextended#1\fi}

\theoremstyle{thmstyleone}%
\theoremstyle{thmstyletwo}%

\theoremstyle{thmstylethree}%

\begin{document}

\title[Article Title]{Early Detection of Water Stress by Plant Electrophysiology: Machine Learning for Irrigation Management}


\author*[1]{\fnm{Eduard} \sur{Buss}}\email{eduard.buss@uni-konstanz.de}

\author[1]{\fnm{Till} \sur{Aust}}\email{till.aust@uni-konstanz.de}

\author[1]{\fnm{Heiko} \sur{Hamann}}\email{heiko.hamann@uni-konstanz.de}

\affil*[1]{\orgdiv{Department of Computer and Information Science}, \orgname{University of Konstanz}, \orgaddress{\street{Universitätsstraße 10}, \city{Konstanz}, \postcode{78464}, \state{Baden-Württemberg}, \country{Germany}}}

 \abstract{\textbf{Purpose:} Fast detection of plant stress is key to plant phenotyping, precision agriculture, and automated crop management. In particular, efficient irrigation management requires early identification of water stress to optimize resource use while maintaining crop performance. Direct physiological sensing offers the potential to detect stress responses before visible symptoms appear.
 \textbf{Methods:} In this study, we recorded electrophysiological signals from greenhouse-grown tomato plants subjected to water stress and developed a framework based on machine learning for online stress detection. The recorded time-series data were processed using a processing pipeline that includes statistical feature extraction and selection, automated machine learning or alternatively deep learning, and probability calibration. 
 \textbf{Results:} Across multiple input time horizons, we found that a 30-minute look-back window strikes the best balance between rapid decision-making and classification performance. Using automated machine learning, the framework achieved classification accuracies of up to 92\%, outperforming deep learning approaches. Sequential backward selection reduced the feature set while maintaining performance. Importantly, the framework detects transitions from healthy to stressed states in recordings that were not included in the training set.
\textbf{Conclusion:} Overall, we provide a decision-support tool for farmers and establish a foundation for biofeedback-driven irrigation control to improve resource efficiency in (semi-)autonomous crop production systems.}

\keywords{Precision Agriculture, Resource Management, Automated Machine Learning, Deep Learning, Decision Support System, Plant Electrophysiology}



\maketitle

\section{Introduction}\label{sec1}

The European Environment Agency reports that 31\% of total water abstraction in Europe is attributed to agriculture, primarily driven by irrigation~\citep{EEA2025WaterAbstraction}. As water demand continues to increase due to demographic growth, economic activities, and climate change, agriculture remains dependent on water availability to ensure yield stability, crop quality, and farmer resilience against irregular rainfall patterns~\citep{ECA2021WaterUse}. 
Agriculturally used water is sourced from surface water bodies such as rivers and lakes, groundwater reserves, rainwater harvesting, and reclaimed wastewater. However, its use in agriculture directly influences overall water quantity and quality through pollution from fertilizers and pesticides~\citep{thompson2020reducing}. The European Common Agricultural Policy (CAP), as the primary framework governing agriculture and rural development, promotes more efficient water use to mitigate water scarcity~\citep{EC2025_CAP28Implementation}. 
Sustainable and efficient management of natural resources constitutes a key strategic objective of CAP. 
Traditional irrigation practices, such as commercial pre-programmed irrigation controllers, distribute the same amount of water over the whole field, ignoring local differences in soil conditions and the plants' real hydration requirements~\citep{abioye2020review}.

There has been an increased focus on the development of advanced irrigation systems in recent decades due to the growing challenges posed by water scarcity, climatic variability, and the need for efficient resource use. Key advances have been in irrigation monitoring, for example, via the Internet of Things (IoT), and control systems~\citep{abioye2020review}. 
These efforts have established precision irrigation: data-driven irrigation strategies that modulate water delivery optimized in quantity and timing, adapted to the water requirements of crops across varying field conditions and growth stages to reduce water consumption~\citep{abioye2020review}.

Agricultural sensor networks should not only monitor environmental conditions but also directly assess plant physiological status, as environmental variables are only indirect predictors of plant stress. This concept has been described as the Internet of Plants (IoP)~\citep{steeneken2023sensors}, where networks of sensors continuously monitor plant health and detect both biotic and abiotic stress. Similar to human health-monitoring technologies such as wearable sensors, dedicated devices are emerging that directly measure plant physiological signals~\citep{ataei2023internet}. Typical sensing modalities include sap flow sensors, stem diameter sensors, multispectral imaging, and electrophysiological recordings. 
To achieve field-scale coverage with high spatial resolution, these technologies adopt concepts from the IoT. They typically operate autonomously, communicate wirelessly, rely on low-power electronics, and may incorporate energy-harvesting modules such as solar power. Although IoP systems have strong potential to provide actionable insights and to enable \mbox{(semi-)autonomous} crop management. However, there are open questions about which physiological parameters to monitor and which sensing technologies are most suitable. 


Plants, as sessile organisms, must continuously adapt to complex and changing environmental conditions~\citep{johns2021fast}. They rely on sophisticated sensing systems to perceive environmental cues and transmit locally detected information throughout the plant to coordinate systemic responses. A~key class of internal signals is ion fluxes at the cellular level, including Ca\textsuperscript{2+}, K\textsuperscript{+}, Cl\textsuperscript{–}, and H\textsuperscript{+}, which generate electrical signals that propagate through the plant. One example is the action potential, a short transient change in membrane potential. In species such as \textit{Dionaea muscipula} (Venus flytrap) and \textit{Mimosa pudica}, these signals trigger rapid leaf movements for prey capture or surface reduction~\citep{fromm2007electrical}.
 Despite their significance, measuring and analyzing plant electrical signals under real-world constraints remains challenging. Recording is typically performed either invasively using metal electrodes~\citep{li2021plant}, which cause tissue damage, or non-invasively using surface electrodes~\citep{meder2021ultraconformable}, which are dependent on surface moisture and are affected by plant growth and mechanical movement (e.g., wind). 
Signal analysis outside controlled laboratory settings is further complicated by multiple stimuli that simultaneously trigger overlapping electrical signals~\citep{steeneken2023sensors,li2021plant}. 
Plants also differ in physiological state, morphology (e.g., number of leaves, height), and developmental stage, leading to inter-individual variability in recorded signals~\citep{huber2016long}. 
These factors require advanced analytical approaches, including machine learning, to reliably extract meaningful patterns from complex and heterogeneous data.

Recent studies have explored the use of machine learning methods to analyze plant electrophysiological signals during abiotic stress like drought \citep{najdenovska2021identifying,tran2019electrophysiological,zhou2025machine}, salinity~\citep{bhadra2023multiclass,zhou2025machine}, temperature \citep{Aust2025a,buss2023stimulus}, ozone~\citep{bhadra2023multiclass,Aust2025a}, nutrient deficiencies~\citep{gonzalez2023detecting,najdenovska2021identifying} as well as biotic stresses like spider mites \citep{najdenovska2021identifying} or caterpillars~\citep{reissig2021fruit}. 
For instance, \citet{najdenovska2021identifying} recorded electrophysiological signals from 36~tomato plants exposed to drought, nutrient deficiencies, and spider-mite infestation. Electrical activity was measured using PhytlSigns devices (Vivent SA) with electrodes inserted into the stem. From the signals, 34~statistical features were extracted across multiple temporal windows, and an XGBoost classifier achieved accuracies of up to 85\% for a 1~min window.
In this work, we extend their approach in two ways. First, we extract a much larger set of statistical descriptors using the \textit{tsfresh} framework~\citep{Christ2018tsfresh}. Second, we use automated machine learning (AutoML) to optimize the preprocessing and classification pipeline instead of manually selecting a classifier. Furthermore, rather than completely stopping irrigation, we study graded irrigation regimes with varying water amounts to better reflect realistic agricultural conditions.

\citet{gonzalez2023detecting} studied electrical activity in 16~tomato plants under nitrogen deficit in a greenhouse using PhytlSigns sensors (Vivent SA). Plants were grown in coconut fiber and monitored for 15~days, initially under a standard nutrient solution (normal state) and later with nitrogen reduced to one-third (stressed state). Four deep learning architectures were evaluated on time series windows of 1~s to 30~s, with the encoder model achieving up to 99\% accuracy. Classifier certainty over the full measurement period revealed a transition from the normal to the stressed state.
Similarly, we aim to detect the transition from healthy to stressed plants using classifier certainty. In contrast, we evaluate model generalizability using separate training, validation, and test sets. Because we focus on drought stress, we include control plants maintained under optimal conditions to account for environmental fluctuations in semi-controlled greenhouses. To address potential bias in classifier certainty, we apply probability calibration. While both studies use PhytlSigns sensors, we additionally validate this sensing approach using our own system, \PhytoNode.

\Aust~\textit{et~al.}~\citep{Aust2025a} also employed the \PhytoNode\ to investigate elevated temperature and ozone levels based on the electrophysiological response of ivy (\textit{Hedera helix}) plants. They trained a fully convolutional network~(FCN) on the recorded signals and deployed the trained model directly on the microcontroller-based \PhytoNode. Validation accuracies of up to 86\% and 88\% were achieved for temperature and ozone stimuli, respectively. The network was transferred to the microcontroller using their custom toolchain, Mbed Torch Fusion~OS, which enables direct deployment of PyTorch models onto the \PhytoNode. Subsequently, they successfully detected increased temperature and ozone levels in online classification experiments.

Discrete predictions, such as `healthy' or `stressed', alone are insufficient for practical decision-making. Prediction confidence is equally important, as incorrect actions directly affect resource use and economic outcomes. For example, a stress prediction with only 52\% confidence may require human review or higher thresholds for automated interventions. Depending on the classifier and training procedure, predicted probabilities can be systematically miscalibrated~\citep{guo2017calibration}. Neural networks trained with cross-entropy often become overconfident, while ensemble methods such as random forests may yield underconfident estimates. 
Guo~\textit{et}~al.~\citep{guo2017calibration} evaluated six calibration methods, including isotonic regression, Platt scaling, temperature scaling, histogram binning, Bayesian binning, vector scaling, and matrix scaling for neural network classifiers and identified temperature scaling as a simple and effective approach.

We contribute to the concept of the Internet of Plants through the deployment of our self-sustaining sensor node, called \PhytoNode~\citep{buss2025phytonode}, that communicates wirelessly and harvests solar energy. Our sensor node measures plant electrophysiological signals, and we have validated it successfully in outdoor experimental setups~\citep{buss2025plants}. 
In this work, we extend the application from environmental monitoring to water-stress detection for precision irrigation. We include the plant physiological state in the irrigation system and form a bio-hybrid system of plant, \PhytoNode, and machine-learning-based analysis. An approach that can function as a decision-support tool for farmers. 

\section{Materials and Methods} \label{Sec. MaterialsAndMethods}
\begin{figure}[t] \centering
\includegraphics[width=12 cm]{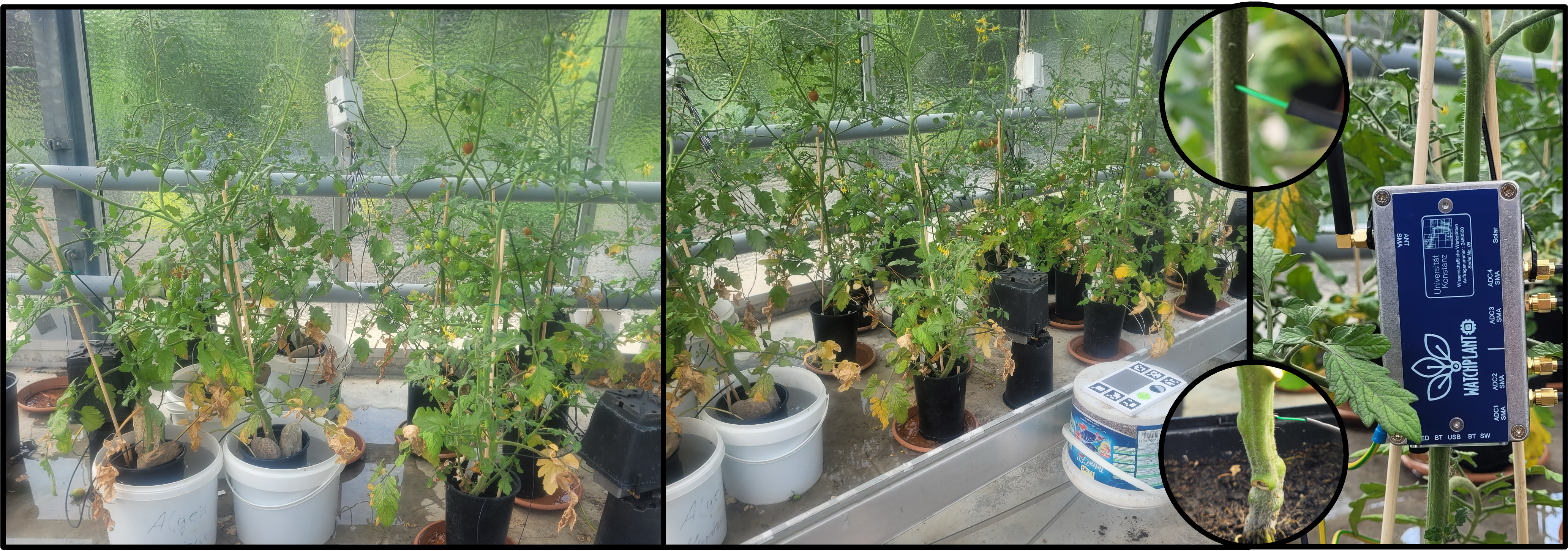}
\caption{Experimental Setup. Left: 4~plants under saturated conditions and 4~plants receiving 400~ml (June 8). Middle: All 16~plants. Right: Our \PhytoNode\ attached to a tomato plant in a greenhouse with a close-up of the silver electrodes.} \label{fig experimental setup}
\end{figure}

We measured the electrical potential of tomato plants using our low-power, self-sustaining sensor node, called \PhytoNode\ (see Fig. \ref{fig experimental setup}, right)~\citep{buss2025phytonode}. It is enclosed in a weather-resistant housing and includes a~4,800~mAh LiPo battery that is recharged by a solar panel that delivers up to 5~W under greenhouse lighting conditions. Each \PhytoNode\ can simultaneously record electrical signals from two plants and transmit the data via Bluetooth Low Energy (BLE) to a data sink, a~Raspberry Pi~4 in this study. Electric differential potentials (EDP) per plant were acquired using two silver-coated electrodes inserted into the stem, one near the base and the other at least 30~cm higher along the stem. The signals were sampled at 10~Hz.

We recorded the electrical differential potential of 16~tomato plants continuously over a period of 18~days, from June 4th to June 22nd, 2025 (see Fig.~\ref{fig experimental setup}, left and middle). During the initial phase (June 4th to 8th), all plants were irrigated daily with 400~mL of water, as this amount resulted in a dry or minimally moist soil surface the following day. After this period, the plants were divided into four groups, each subjected to a distinct irrigation regime for the remainder of the experiment duration: four plants were placed in a bucket with constant water access (saturated condition), four plants continued receiving 400~mL per day (ideal condition), four plants received 200~mL per day (moderate water deficit), and the remaining four plants received 100~mL per day (strong water deficit). In addition to the electrical potential, soil moisture was monitored at 0.1~Hz using standard capacitive soil moisture sensors. All recorded and processed data are available online~\citep{buss2026_zenodo}.

\section{Machine Learning Analysis} \label{Sec: Machine Learning Analysis}
We employ two distinct machine learning pipelines: (1)~a feature-based AutoML pipeline (Sec.~\ref{Sec. methods FeatureBasedModeling}) and (2) an end-to-end deep learning pipeline with hyperparameter optimization (Sec.~\ref{Sec. methods DeepLearningHPO}). The former comprises (1a)~preprocessing of raw measurements, time-window slicing and class label assignment, (1b)~statistical feature extraction for each time window, (1c)~AutoML to identify suitable classifier pipelines, and (1d)~sequential backward selection to determine a minimal feature subset.
The second pipeline includes (2a)~the same preprocessing step as in~(1a), extended by robust z-score normalization, (2b)~hyperparameter optimization of three neural network architectures, and (2c)~extended training of the best-performing configurations.
Finally, we apply class certainty calibration (Sec.~\ref{Sec. Methods Confidence Calibrationi}) to the resulting classifiers to mitigate internal biases and analyze the temporal characteristics of the predicted confidence scores on previously unseen data (Sec.~\ref{Sec. temporal transisitons}).

\emph{(1a \& 2a) Preprocessing, Time Window Slicing and Label Assignment.}
We resampled the electrical potential to 1~Hz, as this smoothed the data without significant distortion. 
Subsequently, all time series were sliced into fixed time windows of 1~min, 5~min, 30~min, 1~h, and 6~h.

The dataset was labeled using two approaches, following a strategy similar to that described in~\citep{gonzalez2023detecting}. In both cases, measurements from the initial days were assigned to the healthy class, whereas measurements from the final days were labeled according to the specific classification scheme adopted in each approach.

In the first approach, a binary classification task (healthy vs. stressed) was defined. The final days of the treatment groups overwatered, 200~ml, and 100~ml were labeled as `stressed', irrespective of the specific irrigation regime, while the initial days across these groups were labeled as `healthy'.
In the second approach, a multiclass classification framework was employed to distinguish between overwatered, underwatered, and healthy conditions. Measurements from the final days of the overwatered group were assigned to the `overwatered' class, while measurements from the final days of the four plants in the 100 ml group were labeled as `underwatered'. The initial days of both groups were used to define the `healthy' class.

For both approaches, intermediate days were not used for training but instead served to evaluate the temporal behavior of the classifier, specifically to identify the point at which predictions transition from healthy to stressed conditions. Our analysis primarily focuses on the binary classification approach, since the multiclass setting exhibited strong validation performance but limited generalization to unseen test data (see Sec. \ref{Sec. Res. Optuna and DeepLearning} for details).

\subsection{Feature-Based Modeling and AutoML} \label{Sec. methods FeatureBasedModeling}
\emph{(1b) Feature Extraction.}
For each time window, we extracted approximately 700~time series characteristics, also called features (see Tab.~\ref{Tab. datasets}) using the Python library \textit{tsfresh}~\citep{Christ2018tsfresh}. Feature values resulting in numerical overflows (±inf) were replaced by the corresponding feature mean. Subsequently, features with variance below 0.01 were removed, as such features provide limited discriminative power. 
For numerical stability and feature comparability, features are normalized using min–max scaling given by $x_{\text{i, norm}} = \frac{x_i - \min(\textbf{x})}{\max(\textbf{x}) - \min(\textbf{x})}$ where $x_i$ represents the current sample, and $\textbf{x}$ the entire set of one feature. 

In addition to the four plants that received 400~ml over the entire period, we excluded two further plants (one each from the overwatered and 100~ml groups) from the training process and used them as test plants. The remaining data was split into training and validation sets using an 80/20 ratio. The split was stratified to preserve the class and plant data distributions across both subsets.

\emph{(1c) AutoML.}
We identified a potential classification pipeline using the AutoML framework \textit{Naive AutoML}~\citep{Mohr2022}. AutoML aims to identify suitable preprocessing and learning algorithms that achieve strong generalization performance on a given dataset. This involves both selecting an appropriate pipeline and determining an optimal set of hyperparameters. First, it searches for a pipeline composed of preprocessing steps and a single predictor implemented in the Python framework \textit{sklearn}. During this stage, candidate pipelines are evaluated using their default hyperparameter settings. In a second optimization phase, the best-performing pipeline is selected, and its hyperparameters are further optimized via random search.
The pipeline search was done using all computed features, allowing up to 100~hyperparameter optimization iterations, and we used accuracy as evaluation metric. 

\emph{(1d) Feature Selection with Sequential Backward Selection.}
Feature selection was performed to identify a compact subset of discriminative features while avoiding unnecessary dimensionality, redundancy, and noise that may degrade classification performance. We applied a staged approach by first using the Mutual information (MI) as a univariate preselection criterion to reduce the dimensionality of the high-dimensional feature space generated by \textit{tsfresh} \citep{vergara2014mutualinformation}. MI quantifies the statistical dependency between an individual feature and the class labels and was computed for all extracted features. The features were ranked in descending order of MI, and the top 200~features were retained as candidate features.

The second stage includes multivariate feature selection using sequential backward selection (SBS) as implemented in the \textit{mlxtend} Python framework~\citep{raschkas_2018_mlxtend}. SBS was applied to the candidate feature set to identify feature subsets with strong joint discriminative power. The algorithm was initialized with the full set of candidate features and iteratively removed one feature at a time. At each iteration, all possible feature subsets obtained by excluding a single feature were evaluated, and the subset yielding the highest classification performance was retained. This process was repeated until only one feature remained. 
Classification performance during feature selection was assessed using accuracy, which is appropriate given the balanced datasets. Furthermore, performance is assessed using five Group K-Fold splits, in which the classifier is trained five times, and in each split, two plants are excluded from training and used for validation.

\subsection{Deep Learning Models and Hyperparameter Optimization}\label{Sec. methods DeepLearningHPO}
\emph{(2a) Robust Z-Score Normalization.} Compared to classical machine learning approaches, deep learning models are capable of learning feature representations directly from raw measurements. To provide robust normalization that is resilient to outliers, we apply a commonly used robust z-score transformation based on the median and interquartile range (IQR). The normalized value is defined as $\textbf{z} = \frac{\textbf{x} - \text{median}(\textbf{x})} {IQR(\textbf{x})}$ where $\textbf{z}$ denotes the normalized time series, $\textbf{x}$ the raw observation, $\text{median}(\textbf{x})$ the median of the time series, and $IQR = Q_3-Q_1$ represents the interquartile range, i.e., the spread of the central 50\% of the data. This transformation normalizes electrical potentials for comparability across plants and experiments, reducing physiological and experimental variations. Simultaneously, it provides a robust alternative to standard z-score normalization by reducing sensitivity to outliers caused by transient artifacts, such as mechanical perturbations during plant handling or sensor-related noise. 

\emph{(2b) Hyperparameter Search with Optuna.} Hyperparameter search is a labor intensive task in machine learning. This section describes the used DL architectures in combination with \textit{Optuna} \citep{akiba2019optuna}. \textit{Optuna} is an open-source framework for hyperparameter optimization that uses sampling algorithms to identify optimal hyperparameter configurations within a predefined search space. We explore this space across 100 trials using the Tree-structured Parzen Estimator (TPE) sampling strategy. The \textit{HyperbandPruner} enables early termination of unpromising trials by employing multiple successive halving procedures. Trials are allocated to different brackets: early brackets evaluate a large number of hyperparameter configurations with a limited training budget (exploration), whereas later brackets focus on fewer configurations with increased training resources (exploitation). Trough recurrent comparisons in a bracket only one third survives and will not get pruned. Each model configuration is trained for up to 100 epochs with early stopping after 15 epochs without improvement of the validation loss. We define the hyperparameter search space of each model in Table~\ref{tab:optuna-hyperparameter}. Next we are describing our fundamental architectures, which are the starting point for the hyperparameter optimization.
\\ 
\\
\noindent\textbf{\textit{Convolutional Neural Network (CNN).}}\\
Our first and simplest model is a standard CNN.
It serves as our baseline deep learning classifier as it has already been used to classify electrical plant signals \citep{gonzalez2023detecting}. 
Our architecture is composed of sequential convolutional blocks, each consisting of a 1D~convolutional layer, batch normalization, and a ReLU activation function. We add a max pooling and dropout layer after each second convolutional block to avoid excessive dropouts and shrinking of the feature space. 
The resulting feature maps are aggregated using an 1D~adaptive average pooling layer to make the architecture independent of the input length followed by a flatten operation. 
This produces a one-dimensional feature vector that is passed to a fully connected artificial neural network (ANN) consisting of one or several (optimized by Optuna) linear layers with ReLU activations. 
With each successive layer, the dimensionality of the feature vector is reduced by half, except for the final layer, which outputs a vector whose length corresponds to the number of classes in the classification task. 
We search for general learning hyperparameters (learning rate, weight decay, batch size) as well as model specific hyperparameters such as number of convolutional blocks, kernel size, channel dimension, dropout rate and the number of linear layers (see Table \ref{tab:optuna-hyperparameter}) using \textit{Optuna} as a optimization framework. 
\\
\\
\noindent\textbf{\textit{InceptionTime.}}\\
The second model, InceptionTime, is an ensemble of CNN models~\citep{ismail2020inceptiontime}. 
We adapt the PyTorch implementation provided by Campos \textit{et al.}~(2023)~\citep{pacmmod/0002Z0KGJ23}. 
The receptive field of each CNN model differs and is defined by its kernel size which enables the network to extract temporal features at different scales. This property makes it well suited for classifying electrical signals that exhibit both short-term dynamics (e.g., ion fluxes) and long-term patterns (e.g., stress accumulation or circadian rhythms). 
Our architecture is composed of multiple sequential Inception modules organized into residual blocks. Each module shares the same structural design but uses different initial weights and applies convolutional filters of multiple lengths in parallel to capture both short-term and long-term temporal patterns. 
The module begins with a bottleneck layer that is a learned channel-mixing layer to reduce the input dimensionality to decrease computational cost. The bottleneck output is then passed to several parallel convolutional layers with different kernel sizes. An additional parallel branch performs max pooling on the raw input, followed by another bottleneck layer, to provide robustness to small perturbations.
All parallel convolutions are first concatenated, then batch-normalized, and activated using a ReLU nonlinearity. 
These Inception modules are arranged sequentially within residual blocks, with skip connections linking the input and output of each block. A global average polling layer complete sthe architecture.
Using Optuna, we perform hyperparameter optimization over both, general learning parameters and architectural choices. 
This includes the number of residual blocks, the number of Inception modules per block, the number of parallel convolutions per module, kernel sizes, bottleneck channel counts, and the number of output channels in each module (Table~\ref{tab:optuna-hyperparameter}). 
Importantly, we only optimize the output channel size of the first block, as it is doubled in each subsequent block. 
Similarly, we specify the maximum kernel size, which is reduced by 20\% for each further parallel convolution.
\\
\\
\noindent\textbf{\textit{Mamba.}}\\
Our next architecture is the selective State Space Model (SSM) known as Mamba \citep{gu2024mamba}. 
This sequential model integrates principles from recurrent neural networks, convolutional neural networks, and classical continuous-time state space models. A standard state space system is expressed as~$h^{\prime}(t)=A h(t)+ B x(t), y(t)=C h(t)+D x(t)$ a formulation widely used in control theory \citep{gu2021efficiently}. 
Here,~$h(t)$ denotes the system’s hidden state,~$x(t)$ the current input, and~$y(t)$ the output. The transition matrix A control the change of the hidden state, while the input matrix B determines how new input signals influence the updated state~$h^{\prime}$. Likewise, the output matrices~$C$ and~$D$ map the internal state and the input, respectively, to the output~$y(t)$, with~$D$ functioning analogously to a skip connection. We adopt this modeling principle because prior work has demonstrated that SSM architectures can capture long-range temporal dependencies spanning 10,000 time steps or more, while maintaining linear computational complexity with respect to input length~\citep{gu2021efficiently}. 
In contrast, self-attention based architectures typically struggle at such temporal horizons and exhibit quadratic scaling. 
This property makes SSMs particularly well suited for electrophysiological plant measurements, which are frequently sampled at high rates often up to 500~Hz.
Our model is a compact sequence encoder built around Mamba blocks. The network begins with a linear projection layer that maps the raw input to the model's internal dimensionality. The projected sequence is then passed through one or more identical processing layers (mamba layer). Each layer applies layer normalization, followed by a Mamba block as described by Gu~\textit{et al.}~(2024)~\citep{gu2024mamba}, then a dropout layer. 
The output of this layer is combined with the original layer input through a residual connection. After the final mamba layer, the sequence of hidden states is condensed into a single vector using mean pooling across the temporal dimension. This pooled representation provides a global summary of the entire input sequence. A final linear layer maps this representation to the model’s output dimension which is the number of classes in our classification.
We use \textit{Optuna} for automated hyperparameter optimization that covers again general learning parameters and structural design choices. The Mamba related parameters include the dimensionality of the input projection, the hidden size of the model, the kernel size of the initial convolution, and the expansion factor used in the block’s gating mechanism. 
Beyond the Mamba block itself, we search only for the number of stacked Mamba layers (see Table~\ref{tab:optuna-hyperparameter} for details). 

\begin{table}[h]
\centering
\small
\caption{Hyperparameter search space for different model architectures optimized with \textit{Optuna} (log-scaled where indicated).}
\label{tab:optuna-hyperparameter}

\begin{tabular}{lll}
\toprule
Model & Hyperparameter & Range \\
\midrule

\multirow{3}{*}{General}
& Learning Rate & $[10^{-4},\,10^{-2}]_{\log}$ \\
& Weight Decay & $[10^{-6},\,10^{-3}]_{\log}$ \\
& Batch Size & $2^k,\; k \in \{4,5,6,7\}$ \\

\midrule
\multirow{5}{*}{CNN}
& \# Conv. Blocks & $2k,\; k \in \{1,\dots,10\}$ \\
& Kernel Size & $2^k + 1,\; k \in \{2,\dots,7\}$ \\
& Channel Dimension & $2^k,\; k \in \{1,\dots,10\}$ \\
& Dropout & $[0,\,0.5]$ \\
& \# Linear Layers & $\{1,\dots,6\}$ \\

\midrule
\multirow{5}{*}{InceptionTime}
& \# Residual Blocks & $\{2,\dots,6\}$ \\
& \# Inception Modules & $\{1,2,3,4\}$ \\
& \# Parallel Convolutions & $\{2,\dots,6\}$ \\
& Output Channels & $2^k,\; k \in \{3,4,5,6\}$ \\
& Kernel Size & $2^k + 1,\; k \in \{2,\dots,7\}$ \\

\midrule
\multirow{5}{*}{Mamba}
& \# Mamba Layers & $\{1,\dots,8\}$ \\
& Input Dimension & $2^k,\; k \in \{4,\dots,9\}$ \\
& Hidden Size & $2^k,\; k \in \{3,4,5,6\}$ \\
& Kernel Size & $\{2,3,4\}$ \\
& Expansion & $\{1,\dots,5\}$ \\

\bottomrule
\end{tabular}
\end{table}

\subsection{Confidence Calibration with Temperature Scaling}\label{Sec. Methods Confidence Calibrationi}
Temperature scaling rescales the classifier’s logits $\mathbf{z}_i$ by a scalar temperature $T$, producing  $\mathbf{z}_i / T$, before applying a sigmoid or softmax to obtain calibrated probabilities. Temperature scaling is argmax-invariant and does not change classification accuracy as $T$ is optimized after training and applied uniformly to all logits. 
The effect of calibration is visualized using reliability diagrams, which compare empirical class frequencies (based on true labels) with predicted certainties. These diagrams are typically combined with histograms showing the distribution of prediction confidences~\citep{guo2017calibration}. 
Calibration performance is further quantified using two commonly used metrics for binary classification: the Brier score and the Adaptive Calibration Error~(ACE)~\citep{glenn1950verification,nixon2019measuring}. The Brier score is defined as 
\begin{equation}
\text{Brier} = \frac{1}{N} \sum_{i=1}^N\left(y_i-\hat{p}_i\right)^2\;,
\end{equation}
the mean squared error 
between the predicted probabilities~$\hat{p}_i$ and the corresponding true labels~$y_i$ across all samples $N$. 
While the Brier score measures how closely predicted probabilities match the true labels on a per-sample basis, ACE quantifies how well predicted confidences match the empirical label frequencies. To compute ACE, we first bin the predicted confidences using quantile binning with $M=20$, so that each bin $m$ contains approximately 5\% of the samples. For each bin, the empirical accuracy $\text{acc}(m)$ is computed as the fraction of the true labels of the samples assigned to that bin. In addition, the mean predicted confidence $\text{conf}(m)$ is obtained by averaging the predicted certainties of the samples within the same bin. The ACE is then calculated as 
\begin{equation}
\mathrm{ACE}=\frac{1}{M} \sum_{m=1}^M|\operatorname{acc}(m)-\operatorname{conf}(m)|\;,
\end{equation}
for the average absolute difference between \text{acc}(m) and \text{conf}(m) across all bins.

%
%
\section{Results and Discussion}
\begin{figure}[t] \centering
    \input{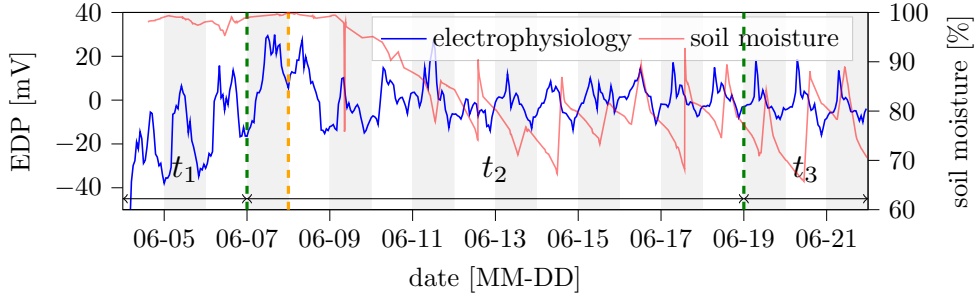}
    \caption{
    Exemplary EDP (blue) and soil moisture (red) of a plant irrigated with 200 mL after day 4 (yellow dashed). Intervals $t_1$, $t_2$, $t_3$ are used for training, validation, and testing. Gray–white shading indicates 24-h cycles.} 
    \label{fig:classification_task}
\end{figure}

We illustrate the classification task using an exemplary measurement of a tomato plant that received 200~mL of water after the fourth day (see Fig.~\ref{fig:classification_task}). As described in Sec.~\ref{Sec. MaterialsAndMethods}, we use the first three days (interval~$t_1$) and label them as `healthy', and we use the last three days (interval~$t_3$) and label them as `stressed'. These data are used to train, validate, and test the classifier.
The interval~$t_2$ in between is analyzed to assess classification certainty over the experiment duration on unseen data and to identify the transition from `healthy' to `stressed'.
We analyze multiple classifier input intervals as shorter time windows enable faster decisions but may contain insufficient information for reliable predictions (see Tab.~\ref{Tab. datasets}).
Naturally, the length of the measured time series increases from 60~data points at a window length of 1~min to 21,600~data points at 6~h intervals. Conversely, in the training set, the number of available time series decreases from 69,112 at 1~min windows to 195 at 6~h windows. The number of extracted features per window length remains approximately constant at around 670~features, except for the 1~min interval, which retains only 421~features per time series. This reduction results from the shorter time-series length and the preprocessing step that removes features with variance below 0.01.

\begin{table} [h]
\setlength{\tabcolsep}{2.7pt}

\centering
\begin{tabular}{cccccccccccccccc}
\toprule
\multirow{2}{*}{data set} 
& \multicolumn{3}{c}{1 min}
& \multicolumn{3}{c}{5 min}
& \multicolumn{3}{c}{30 min}
& \multicolumn{3}{c}{1 h}
& \multicolumn{3}{c}{6 h} \\
\cmidrule(lr){2-16}
& $l$ & $s$ & $f$
& $l$ & $s$ & $f$
& $l$ & $s$ & $f$
& $l$ & $s$ & $f$
& $l$ & $s$ & $f$\\
\midrule
training
& 60    & 69,112   & 421
& 300   & 13,822    & 685
& 1,800 & 2,302     & 669
& 3,600 & 1,158     & 664
& 21,600&  195      & 653\\
validation
& 60    & 17,278    & 421
& 300   & 3,458     & 685
& 1,800 & 578       & 669
& 3,600 & 282       & 664
& 21,600& 45        & 653   \\
test
& 60    & 17,278    & 421
& 300   & 3,456     & 685
& 1,800 & 576       & 669
& 3,600 & 288       & 664
& 21,600& 48        & 653  \\
\bottomrule
\end{tabular}
\caption{All data sets of different timeseries length $l$, data set size $s$, and number of features $f$ based on varying look-back horizons.}
\label{Tab. datasets}
\end{table}

\subsection{Automated Machine Learning and Feature Selection}

Since identifying an appropriate classification pipeline, including preprocessing steps, a classifier, and a well-tuned set of hyperparameters, requires substantial effort, we employed NaiveAutoML on the datasets listed in Table~\ref{Tab. datasets}. 
NaiveAutoML selected Histogram Gradient Boosting~(HGB) as the best-performing classifier for all window lengths except for the 1~min interval. The Extra Trees Classifier (ETC) and Random Forest Classifier were excluded from this window because they yielded only a marginal performance gain of less than 1\% and would require a different probability calibration approach, which compromises methodological consistency and comparability across time windows. HGB is an ensemble method that sequentially builds decision trees, where each tree is trained to correct the residual errors of the previous ensemble. Unlike classical gradient boosting, HGB discretizes continuous features into histograms prior to training, improving computational efficiency.
Furthermore, the NaiveAutoML pipeline applied an additional variance-threshold preprocessing step for the 1~h window lengths to remove features with zero variance. Although features with variance below 0.1 had already been excluded during prior preprocessing, NaiveAutoML may re-identify constant features due to its internal K-fold cross-validation procedure~\citep{Mohr2022}. Within individual folds, feature variance can decrease to zero, leading to their removal.
All investigated pipelines achieved 100\% training accuracy, indicating that the models have sufficient capacity to fully fit the training data (Table~\ref{Tab. NaiveAutoML_performance}). Validation accuracy decreases with increasing look-back horizon, whereas test accuracy increases. For short windows, samples are temporally closer and therefore more similar, which can inflate validation accuracy without reflecting true generalization. Larger windows capture more temporal variability, making validation more challenging but improving generalization on the test set.


\begin{table}[t]
\centering
\begin{tabular}{c|c|ccc||c|ccc}
\toprule
\multirow{2}{*}{horizon} & 
\multirow{2}{*}{pipeline} & 
\multicolumn{3}{c||}{AutoML acc. [\%]} &
\multirow{2}{*}{\# features} &
\multicolumn{3}{c}{SBS acc. [\%]}\\

 &   & train & val. & test  &   & train & val. & test \\
\midrule
1 min  & HGB      & 100 & 92.2 & 62.6 & 96 & 100 & 89.6 & 61.6\\ 
5 min  & HGB      & 100 & 92.6 & 76.3 & 193& 100 & 92.1 & 75.6\\ 
30 min & HGB      & 100 & 91.0 & 83.2 & 182& 100 & 91.4 & 82.5\\ 
1 h    & VT + HGB & 100 & 90.1 & 84.0 & 200& 100 & 90.1 & 82.3\\ 
6 h    & HGB      & 100 & 77.8 & 89.6 & 15& 100 & 82.2 & 87.5 \\ 
\bottomrule
\end{tabular}
\caption{AutoML-determined optimal classification performance using all features (AutoML acc.) and classification performance after sequential backward selection (SBS acc.).}
\label{Tab. NaiveAutoML_performance}
\end{table}

Our feature selection pipeline first ranks features by MI and then identifies an optimal subset using SBS. Initially, the top 200~features were considered. For the 1~min horizon, this was reduced to 100 due to the large training set (69,112~samples) and the combinatorial complexity of SBS, reducing the required model fits from 20,099 to 5,049. Despite this reduction, validation accuracy decreased by only 2.6\% using 96~selected features (see Tab.~\ref{Tab. NaiveAutoML_performance}). \ext{Figure~\ref{fig:feature_selection} further indicates that near-maximum performance is achieved with a comparatively small subset of features.}
For the 5~min, 30~min, and 1~h horizons, SBS selects 193, 182, and 200 features, respectively. In all cases, training, validation, and test accuracies remain nearly identical to those obtained with the full feature set, with the largest decreases of 0.5\% in validation accuracy (5~min) and 1.7\% in test accuracy (1~h). Thus, up to 479 features (5~min) can be removed without meaningful performance loss. For the 6~h horizon, the highest validation accuracy (82.2\%) is obtained using only 15~features, with a corresponding test accuracy of 87.5\% (highest over all horizons). However, due to the smaller dataset size at this horizon, classifiers are more sensitive to individual samples and should therefore be interpreted cautiously.

\ext{Figure \ref{fig:feature_selection} presents the classification accuracy as a function of the number of selected features for the 1~min window length. This horizon is shown because it comprises a smaller maximum feature subset and demonstrates behavior representative of the other short time windows. The 6~h horizon is an exception, as it exhibits markedly more irregular accuracy curves, which can be attributed to the smaller number of available samples.

The training accuracy reaches 99\% with only six features and increases to 100\% by 36~features. The validation accuracy is 84.6\% at six features and increases only gradually as more features are added. Increments of approximately 1\% are observed at 7, 30, 51, and 66~features, leading to a plateau around 89\%. The maximum validation accuracy of 89.6\% is obtained with 96 features.\ebq{should I keep the test accuracy curve?}
Considering only the accuracy values (without certainties), most of the discriminative information appears to be captured by the first few features. For instance, expanding the feature set from 6 to 96 features results in an absolute improvement of 5\% in validation accuracy. Thus, the performance gain associated with larger feature sets is comparatively small relative to the increase in dimensionality.
For deployment on microcontroller-based sensor nodes, where computational and memory resources are limited, a reduced feature set may therefore represent a reasonable trade-off between classification performance and computational efficiency.
\begin{figure}[h]
  \centering
  \input{Images/figure_tikz.tex}
  \caption{Accuracies over the number of features using the 1~min datasets.}
  \label{fig:feature_selection}
\end{figure}
}

\subsection{Optuna and Deep Learning} \label{Sec. Res. Optuna and DeepLearning}
Similarly to NaiveAutoML, we employ Optuna to identify suitable hyperparameter configurations for the selected models (CNN, InceptionTime, and Mamba). The full optimization procedure is documented and publicly available \citep{buss2026_zenodo}. To avoid partially trained models, the best configurations were retrained with an increased number of epochs (200) to ensure full convergence. We conducted multiple independent runs with randomly selected seeds (42, 123, 236, 679, 999) for dataset and model initialization. The resulting mean accuracies and standard deviations are reported in Table~\ref{tab DL results}.
No single model consistently outperforms the others across all horizons. Performance is horizon-dependent: at 1~min, CNN achieves the best result ($63.56\% \pm 1.16\%$); at 5~min, InceptionTime achieves the best result ($70.99\% \pm 1.62\%$); at 30~min, Mamba performs best ($80.21\% \pm 3.91\%$); and at 1~h and 6~h, CNN achieves the highest accuracies ($84.77\% \pm 2.86\%$ and $96.96\% \pm 1.06\%$, respectively).
Overall, CNN provides the best average performance across horizons~(77.57\%) and exhibits comparatively high robustness, with a mean standard deviation of 1.98\% across runs. InceptionTime is competitive~(73.96\%) but shows higher variability~(3.17\%), while Mamba performs worse on average~(70.68\%) and demonstrates the highest variability~(4.32\%), particularly at longer horizons.

\begin{table}[t] 
\setlength{\tabcolsep}{4pt}
\centering
\begin{tabular}{lccc ccc}
\toprule
\multirow{2}{*}{Horizon} 
& \multicolumn{3}{c}{CNN} 
& \multicolumn{3}{c}{InceptionTime} \\
\cmidrule(lr){2-4} \cmidrule(lr){5-7}
& train [\%] & val. [\%] & test [\%]& train [\%] & val. [\%] & test [\%]\\
\midrule
1 min  & 79.45 $\pm$ 1.55 & 71.02 $\pm$ 0.81 & 63.56 $\pm$ 1.16 
       & 74.82 $\pm$ 2.79 & 70.81 $\pm$ 0.53 & 62.25 $\pm$ 1.25\\
5 min  & 80.49 $\pm$ 1.05 & 76.07 $\pm$ 0.63 & 68.97 $\pm$ 1.97 
       & 81.07 $\pm$ 2.11 & 76.84 $\pm$ 0.60 & 70.99 $\pm$ 1.62\\
30 min & 84.31 $\pm$ 1.04 & 76.93 $\pm$ 0.75 & 73.58 $\pm$ 2.83    
       & 89.00 $\pm$ 1.31 & 82.58 $\pm$ 0.28 & 71.06 $\pm$ 2.38 \\ 
1 h    & 77.05 $\pm$ 1.86 & 75.86 $\pm$ 0.56 & 84.77 $\pm$ 2.86    
       & 89.21 $\pm$ 4.78 & 80.56 $\pm$ 2.23 & 77.26 $\pm$ 5.51\\  
6 h    & 83.15 $\pm$ 1.03 & 70.00 $\pm$ 2.54 & 96.96 $\pm$ 1.06 
       & 80.26 $\pm$ 2.86 & 70.87 $\pm$ 6.24 & 88.26 $\pm$ 5.07 \\
\bottomrule

\multicolumn{7}{c}{
\begin{tabular}{lccc}
\multirow{2}{*}{Horizon} & \multicolumn{3}{c}{Mamba} \\
\cmidrule(lr){2-4}
& train [\%] & val. [\%] & test [\%]\\
\midrule
1 min  & 72.66 $\pm$ 0.47 & 70.55 $\pm$ 0.15 & 62.29 $\pm$ 0.79\\
5 min  & 83.62 $\pm$ 3.27 & 76.19 $\pm$ 0.72 & 68.84 $\pm$ 2.01\\
30 min & 81.52 $\pm$ 1.99 & 78.12 $\pm$ 1.29 & 80.21 $\pm$ 3.91 \\ 
1 h    & 89.02 $\pm$ 3.66 & 81.33 $\pm$ 1.84 & 83.79 $\pm$ 2.96 \\ 
6 h    & 67.07 $\pm$ 7.80 & 63.91 $\pm$ 6.82 & 58.26 $\pm$ 11.94 \\
\bottomrule
\end{tabular}
} \\
\end{tabular}
\caption{DL accuracies for the training, validation, and test sets across all look-back horizons, computed over five independent runs with different random seeds.}
\label{tab DL results}
\end{table}

DL models generally perform worse compared to the HGB baseline. For the training accuracy, HGB consistently achieves~100\%, whereas DL models show lower performance, with gaps ranging from about~-11\% (e.g., InceptionTime at~30~min) to~-33\% (Mamba at~6~h). On the validation set, these differences decrease, ranging from~-9\% (Mamba at~1~h) to~-19\% (Mamba at 6~h).
On the test set, the performance gap narrows further and becomes strongly dependent on the prediction horizon. On average, DL models achieve comparable performance to HGB, with a slight mean improvement of approximately $+1\% \pm 1\%$, primarily driven by CNN~(+2\%). At short horizons (5~min and 30~min), DL models consistently underperform, with decreases of~$-6\% \pm 1\%$ and $-8\% \pm 5\%$, respectively. At 1~h, CNN and Mamba outperform HGB by about +2\% and +1\%, respectively. At 6~h, CNN shows the largest improvement (+9\%), followed by InceptionTime (+1\%), whereas Mamba performs substantially worse with a difference of -29\%.
Accordingly, despite partially competitive performance on the test set, we do not further pursue DL approaches for the binary classification task, as HGB provides more consistent and reliable performance across all horizons.
\\
\ext{
\subsection{Multi-class classificaiton}
\ebq{Here are some results of the 3 class case (have not tested every setting), non of the approaches HGB or DL can distinguish between overwatered and underwaterd on the test data. I actually don't want to give it so much space with the table below (if you think it's worth, we can also include it), but I also don't really know how to include this negative results smartly in the paper.}
\begin{table}[h]
\setlength{\tabcolsep}{4pt}
\centering
\begin{tabular}{c c ccc ccc ccc ccc}
\toprule
 &  \multicolumn{4}{c}{AutoML} & \multicolumn{3}{c}{CNN} & \multicolumn{3}{c}{InceptionTime} & \multicolumn{3}{c}{Mamba} \\
\cmidrule(lr){2-5} \cmidrule(lr){6-8} \cmidrule(lr){9-11} \cmidrule(lr){12-14}
Horizon & Pipeline 
& train & val. & test 
& train & val. & test 
& train & val. & test 
& train & val. & test \\
\midrule

1 min  &  &  &  &  & 80.71 & 72.74 & 41.69 &  &  &  &  &  &  \\
5 min  &  & 100 & 94.88 & 51.59 & 83.49 & 78.76 & 40.60 &  &  &  &  &  &  \\
30 min &  & 100 & 94.51 & 47.56 & 87.12 & 84.30 & 48.16 &  &  &  &  &  &  \\
1 h    &  & 100 & 88.69 & 52.77 & 86.54 & 78.94 & 58.59  &  &  &  &  &  &  \\
6 h    &  &  &  &  & 62.72 & 62.72 & 62.72 &  &  &  &  &  &  \\

\bottomrule
\end{tabular}
\caption{Combined table with grouped headers using \texttt{booktabs}.}
\end{table}
}

\noindent\textbf{\textit{Multiclass Classification.}}\\
Both the AutoML and DL approaches were evaluated on a multiclass classification task involving the classes `healthy', `overwatered', and `underwatered' as described in Sec.~\ref{Sec: Machine Learning Analysis}. While the models achieved high performance on the training and validation sets, they consistently failed to generalize to the test data.
For example, with this 30-min interval, the HGB model reached accuracies of 100\%, 95\%, and 48\% on the training, validation, and test sets, respectively. Similarly, the CNN achieved 87\%, 84\%, and 48\% across the same splits. These results were consistent across different intervals and model types.
Overall, the models were not able to reliably distinguish between specific stress types, but rather only between the classes `healthy' and `stressed' plants.

\subsection{Impact of Certainty Calibration}
Certainty calibration adjusts the classifier’s predicted probabilities to better match empirical outcomes, as classifier pipelines can produce under- or overconfident predictions. We therefore apply temperature scaling to the selected model and reduced feature set for all time windows. As expected, classification accuracy remains unchanged because temperature scaling rescales predicted probabilities without altering class labels (Table~\ref{Tab. calibration}). For all horizons we obtain $T>1$, indicating systematic overconfidence. Calibration improves all probabilistic metrics, including negative log-likelihood (NLL), Brier score, and adaptive calibration error (ACE), with the strongest improvement observed for ACE, which directly measures the mismatch between predicted confidence and empirical frequency.

\begin{figure}[t]
    \centering
    \hspace{-0.6 cm}
    \begin{subfigure}[t]{0.4\textwidth}
        \centering
        \input{Images/1min_reliability_curve}
    \end{subfigure}
    \hspace{1.3cm}
    \begin{subfigure}[t]{0.4\textwidth}
        \centering
        \input{Images/1min_confidence_distribution}
    \end{subfigure}
    \caption{(Un)calibrated reliability diagram and histogram data distribution.}
    \label{fig:calibration}
\end{figure}
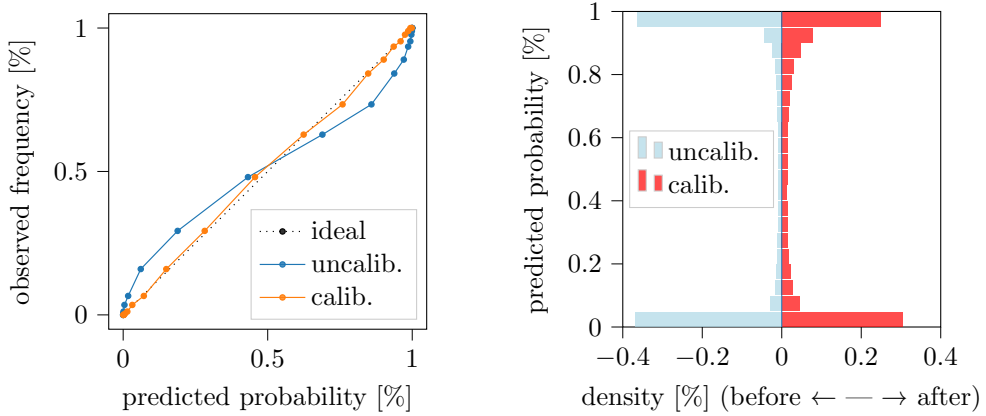

For the 1, 5, and 30~min horizons, calibration reduces ACE to below 1\%, indicating improved probabilistic reliability. In contrast, the 1~h and 6~h horizons retain higher ACE values (2.93\% and 8.13\%), which may result from residual miscalibration not captured by a single global temperature parameter or from increased variance due to the smaller sample size. 
\begin{table}[t]
\centering
\begin{tabular}{c|c|c|cc|cc|cc}
    \toprule
    \multirow{2}{*}{horizon}
    & \multirow{2}{*}{accuracy [\%]}
    & \multirow{2}{*}{$T$} 
    & \multicolumn{2}{c|}{NLL} 
    & \multicolumn{2}{c}{Brier [\%]} 
    & \multicolumn{2}{c}{ACE [\%]} \\
    & & 
    & uncal. & cal. 
    & uncal. & cal.
    & uncal. & cal.\\
    \midrule
    1 min  & 89.6 & 1.58 & 0.2680 & 0.2416 & 7.80 & 7.47 & 4.09 & 0.60\\ 
    5 min  & 92.1 & 3.54 & 0.3991 & 0.1968 & 6.93 & 5.88 & 6.07 & 0.35\\ 
    30 min & 91.3 & 2.97 & 0.3929 & 0.2261 & 7.19 & 6.51 & 5.92 & 0.86\\ 
    1 h    & 90.1 & 1.81 & 0.3130 & 0.2599 & 7.61 & 7.43 & 5.15 & 2.93\\ 
    6 h    & 82.2 & 1.77 & 0.4590 & 0.3958 & 14.00& 12.91& 14.84& 8.19\\
    \bottomrule
\end{tabular}
\caption{Certainty calibration for each interval, including validation accuracy, temperature scaling parameter $T$, and the uncalibrated and calibrated negative log-loss (NLL), Brier score, and adaptive calibration error (ACE)}
\label{Tab. calibration}
\end{table}
To illustrate the overconfident behavior of the classifier and the effect of certainty calibration, we show the reliability diagram and the distribution of predicted probabilities for the 1~min horizon (as a representative example) in Fig.~\ref{fig:calibration}.
Most predictions are concentrated at the extreme probabilities of 0 and 1 (see bins of highest frequency in Fig.~\ref{fig:calibration}, right). 
Temperature scaling shifts probability mass from the extremes toward intermediate confidence levels. For example, 37\% of the samples (6,377 observations) fall into the highest probability bin (100\%) before calibration, whereas this proportion decreases to 30\% (5,249 observations) after calibration. 
The reliability curve is constructed using quantile binning. 
The calibration shifts the curve closer to the ideal diagonal, indicating improved agreement between predicted probabilities and empirical frequencies.

\subsection{Temporal Detection of Class Transitions} \label{Sec. temporal transisitons}
\begin{figure}[t] 
    \centering
    \input{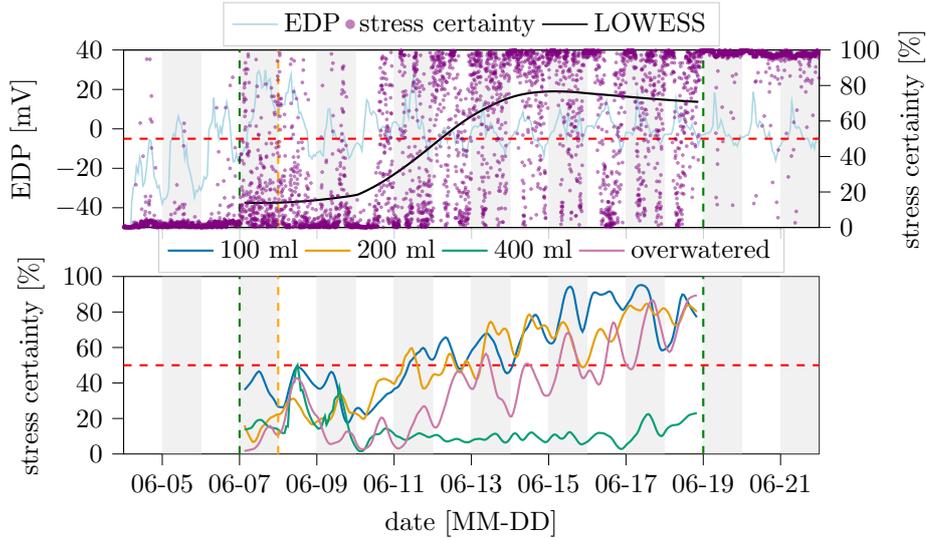}
    \caption{Classification results for the intermediate time interval~$t_2$ (June 7 to June 19, green dashed) lines. Top: Classification outcomes at 5-minute intervals (purple dots) with a 50\% LOWESS-smoothed trend of the certainty values. Bottom: Averaged LOWESS trends (4\% smoothing) for each subgroup.} 
    \label{fig:transition}
\end{figure}

After training the classifiers, the selected models were applied to the full recording period covering the three intervals $t_1$,  $t_2$, and $t_3$ (Fig.~\ref{fig:classification_task}). Fig.~\ref{fig:transition} (top) shows the classification results for the example of a 5-min window of a tomato plant receiving 200~mL of water (same data as in Fig.~\ref{fig:classification_task}). For now, we focus on the purple dots indicating stress certainty in percentages, that is, the classifier's output. Using a 50\% decision threshold, probabilities below 50\% are assigned to the healthy class and values above 50\% to the stressed class.
The certainties indicate a daily pattern: values increase during daytime, shifting toward the stressed class, and decrease at night toward the healthy state. While early measurements are dominated by low certainty, the proportion accumulating near 100\% increases over time, indicating a progressive transition toward the stress state. 

To visualize the dynamic trend, Locally Weighted Scatterplot Smoothing (LOWESS) was applied to the classification certainties. LOWESS is a non-parametric regression method that fits local weighted linear models within a defined neighborhood of data points. To avoid bias from training data, LOWESS was applied only to the central interval~$t_2$. For the upper panel in Fig.~\ref{fig:transition}, a neighborhood size of 50\% of the data was chosen to capture the overall progression of stress development. The resulting smoothed curve crosses the 50\% decision threshold approximately four days after irrigation reduction. 
For the lower panel, LOWESS was first applied to each plant individually, using a smaller neighborhood size of 4\% (corresponding to approximately 12-hour time windows) to resolve daily fluctuations. The four lines shown in Fig.~\ref{fig:transition} are averages of these plant-level LOWESS curves for each treatment group. We observe 24-hour periodicity across all groups, with maxima around midday and minima at night. All reduced-irrigation treatments (100~mL, 200~mL) and the overwatered group show a gradual increase in stress certainty, whereas the control group (400~mL, green line) remains consistently below the decision threshold. The 100~mL and 200~mL groups cross the threshold on day four, whereas the overwatered group reaches it approximately two days later and has generally lower values. The reduced daily amplitude in the control group may reflect more stable physiological dynamics under non-stressed conditions. Given the small group size ($n$=4), these comparisons should be interpreted cautiously due to limited statistical power.


\paragraph{Precision-Recall.}
We analyze the precision–recall (PR) curve for two reasons: (1)~to compare classification performance using the area under the PR curve (AUPRC), and (2)~to adjust classifier behavior by adapting the decision threshold according to farming priorities. The precision is defined as $\frac{t p}{t p+f p} $ and the recall as $\frac{t p}{t p+f n} $ with $tp, fp \text{ and } fn$ being true positives, false positives, and false negatives, respectively. 

With respect to~(1), unlike single-threshold metrics, AUPRC summarizes overall ranking performance across all decision thresholds, providing a threshold-independent measure of class separability. Validation performance remains high (0.91 to 0.976), while test performance increases with longer time windows (Table~\ref{Tab. areaPRcurve}). The 1~min window yields the lowest test AUPRC (0.581), while the 6~h window achieves the highest (0.937). 
\begin{table}[t]
\centering 
\begin{tabular}{c|ccccc}
\toprule
    data set & 1~min & 5~min & 30~min & 1~h & 6~h\\
    \midrule
    validation  & 0.962 & 0.976 & 0.971 & 0.948 & 0.910 \\ 
    test        & 0.581 & 0.728 & 0.871 & 0.871 & 0.937 \\ 
    \bottomrule
\end{tabular}
\caption{Area under the precision–recall curve (AUPRC), all time windows.}
\label{Tab. areaPRcurve}
\end{table}

\begin{figure}[h] \centering
    \input{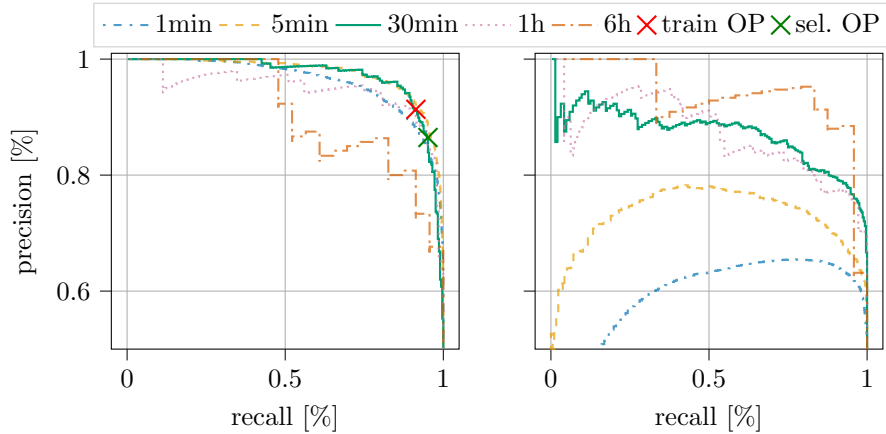}
    \caption{Precision–recall curves of the best classifier of each window applied to the validation (left), and test (right) datasets. The red cross marks the performance of the 30~min classifier at a threshold of 50\%, while the green cross represents the same classifier at a recall of 95\% (threshold of 32\%) to capture more stressed plants at the expense of precision. }
    \label{fig pr_curve}
\end{figure}

With respect to (2), the pipeline can function either as a decision-support tool or as a direct control signal for irrigation. Since priorities may shift (e.g., water conservation vs. growth maximization), the decision threshold can be adjusted using the precision–recall curve (Fig.~\ref{fig pr_curve}).
Treating stressed as the positive class, recall measures the proportion of stressed plants correctly identified, while precision gives the share of true positive stress alarms. Typically, we face a tradeoff here. For example, high precision with low recall reduces false positives (false alarms), but increase false negatives (missed stressed plants), whereas high recall increases detection at the cost of more false positives.
At a 50\% decision threshold, precision and recall are both ~91\% (red cross, Fig.~\ref{fig pr_curve}). Increasing recall to 95\% (green cross) to identify a greater proportion of stressed plants reduces precision to 87\% and increases false positives. This adjustment corresponds to a certainty threshold of 32.1\%. In summary, the PR curve enables systematic tuning of the classifier to application-specific requirements.

\section{Discussion and Conclusion}\label{Sec. Conclusion}
The objective of this work is to develop a decision-support system for irrigation management based on real-time crop physiological measurements. Electrophysiological responses from 16~tomato plants were recorded under different irrigation regimes and analyzed using DL combined with hyperparameter optimization and a machine learning pipeline comprising AutoML and feature selection. Lastly, post hoc certainty calibration was conducted to remove internal classifier biases. 
While deep learning models (especially CNNs) slightly outperform HGB on test data at short (1~min) and long horizons (1~h and 6~h), we chose HGB for further analysis because it provides more consistent performance across all horizons. In addition, its decision-making process can be traced through the individual decision trees, making the classification results more interpretable.
A key challenge was selecting an input time horizon that balances performance and decision latency. Although the 5~min horizon achieved the highest validation accuracy (92.1\%), it performed worse at the test data with only 76\% accuracy compared to 82\% for 30~min window and showed a 0.143 lower test AUPRC (0.871 for 30~min) with a larger validation–test discrepancy of the PR-curves. This indicates reduced reliability, particularly for adaptive thresholding. The 30~min horizon provides a favorable tradeoff between responsiveness and false negative/positive rates and is recommended for implementation. 
Classifier-internal certainty biases can hinder reliable identification of the transition from healthy to stressed states. Temperature scaling corrected the overconfidence of histogram gradient boosting, achieving near-ideal calibration with adaptive calibration errors below 1\%. 
A transition from healthy to stressed states occurred after 4~days in the 100~mL and 200~mL groups and after 6~days in the overwatered group, whereas the 400~mL control group remained stable. This indicates that the electrophysiological changes were driven by irrigation treatments rather than environmental variation. In addition, by excluding two plants from training and testing them independently confirmed generalization to unseen individuals. This suggests that the detected patterns are stimulus-specific rather than individual-specific, despite inherent electrophysiological variability among individual plants.

These findings should be interpreted in light of several limitations. 
The study involved 16~tomato plants under controlled conditions, limiting generalizability to other cultivars, species, seasons, and field environments. Future work will include larger populations and greater environmental variability across seasons. Moreover, only specific irrigation regimes were investigated; additional studies will address other abiotic and biotic stresses, including nutrient deficiencies, diseases, and pests, to broaden applicability.
The default transition point, defined by a 50\% certainty threshold, does not represent definitive physiological stress onset and requires biological validation. This work proposes a data-driven method for detecting the transition from healthy to visibly stressed states. Future studies will integrate plant science expertise and adaptive decision thresholds (see Fig.~\ref{fig pr_curve}) to improve physiological relevance. Another possible approach is to deploy the classifier on independent plant groups using distinct class-specific decision thresholds. Irrigation would be regulated based on model outputs, and performance would be evaluated using independent measures such as multispectral imaging or biomass accumulation.


In summary, we presented a machine learning–guided framework to support irrigation decision-making. Our proposed system can serve as the basis for a biofeedback loop that enables automated irrigation control directly based on plant physiological responses. Such an approach has the potential to improve resource-use efficiency and enhance sustainability in crop production systems.

\backmatter


\section*{Declarations}
\subsection*{Conflict of interest} The authors declare that they have no conflict of interest.
\subsection*{Data availability} Data is available online \citep{buss2026_zenodo}.




\bibliography{mybibliography}

\end{document}

%% file: Images/figure_tikz.tex
\begin{tikzpicture}

\definecolor{darkgray176}{RGB}{176,176,176}
\definecolor{darkorange25512714}{RGB}{255,127,14}
\definecolor{forestgreen4416044}{RGB}{44,160,44}
\definecolor{lightgray204}{RGB}{204,204,204}
\definecolor{steelblue31119180}{RGB}{31,119,180}

\begin{axis}[
height=0.25\linewidth,
legend cell align={left},
legend style={
  fill opacity=0.8,
  draw opacity=1,
  text opacity=1,
  at={(0.97,0.03)},
  anchor=south east,
  draw=lightgray204
},
tick align=outside,
tick pos=left,
width=0.8\linewidth,
x grid style={darkgray176},
xlabel={number of Ffatures},
xmin=-3.95, xmax=104.95,
xtick style={color=black},
y grid style={darkgray176},
ylabel={accuracy [\%]},
ymin=0.386395994906818, ymax=1.02921923833777,
ytick style={color=black}
]
\addplot [semithick, steelblue31119180]
table {%
1 0.644113902071999
2 0.775234402129876
3 0.940285333950689
4 0.976212524597754
5 0.984633638152564
6 0.99412547748582
7 0.994646371107767
8 0.997829609908554
9 0.99855307327237
10 0.99878458154879
11 0.998104525986804
12 0.998726704479685
13 0.998654358143304
14 0.998639888876027
15 0.998524134737817
16 0.99933441370529
17 0.999291005903461
18 0.999551452714435
19 0.999045028359764
20 0.999479106378053
21 0.99972508392175
22 0.999536983447158
23 0.999421229308948
24 0.999479106378053
25 0.999464637110777
26 0.999594860516264
27 0.999667206852645
28 0.999594860516264
29 0.9994501678435
30 0.999681676119921
31 0.999840838059961
32 0.999797430258132
33 0.999985530732724
34 0.999898715129066
35 0.999956592198171
36 1
37 0.999942122930895
38 0.999956592198171
39 0.999971061465447
40 0.999942122930895
41 1
42 0.999971061465447
43 0.999927653663618
44 0.999971061465447
45 1
46 1
47 1
48 1
49 1
50 1
51 0.999985530732724
52 0.999985530732724
53 0.999985530732724
54 1
55 1
56 1
57 1
58 1
59 0.999985530732724
60 0.999985530732724
61 1
62 1
63 1
64 1
65 1
66 1
67 1
68 1
69 1
70 1
71 1
72 1
73 1
74 1
75 1
76 1
77 1
78 1
79 1
80 1
81 1
82 1
83 1
84 1
85 1
86 1
87 1
88 1
89 1
90 1
91 1
92 1
93 1
94 1
95 1
96 1
97 1
98 1
99 1
100 1
};
\addlegendentry{train accuracy}
\addplot [semithick, darkorange25512714]
table {%
1 0.638904965852529
2 0.737006597985878
3 0.816645445074661
4 0.836555156846857
5 0.845583979627272
6 0.846857275147587
7 0.84778330825327
8 0.861963190184049
9 0.862136821391365
10 0.860921402940155
11 0.860053246903577
12 0.86219469846047
13 0.860921402940155
14 0.860400509318208
15 0.86207894432226
16 0.861210788285681
17 0.863699502257206
18 0.861442296562102
19 0.862136821391365
20 0.865088551915731
21 0.862021067253154
22 0.864509781224679
23 0.864278272948258
24 0.863178608635259
25 0.864104641740942
26 0.862136821391365
27 0.865146428984836
28 0.864567658293784
29 0.863178608635259
30 0.865493691399468
31 0.874522514179882
32 0.873307095728672
33 0.87405949762704
34 0.873422849866883
35 0.874348882972566
36 0.874348882972566
37 0.874001620557935
38 0.874464637110777
39 0.873770112281514
40 0.875506424354671
41 0.876027317976618
42 0.873133464521357
43 0.87405949762704
44 0.874175251765251
45 0.874175251765251
46 0.875969440907512
47 0.874291005903461
48 0.876895474013196
49 0.876837596944091
50 0.875159161940039
51 0.875795809700197
52 0.882335918509087
53 0.881988656094455
54 0.882104410232666
55 0.881988656094455
56 0.882335918509087
57 0.882278041439981
58 0.882972566269244
59 0.881699270748929
60 0.882335918509087
61 0.88181502488714
62 0.880541729366825
63 0.88326195161477
64 0.881004745919667
65 0.883030443338349
66 0.883782845236717
67 0.890207199907397
68 0.895300381988656
69 0.892580159740711
70 0.893216807500868
71 0.894721611297604
72 0.894258594744762
73 0.893506192846394
74 0.893911332330131
75 0.894432225952078
76 0.894316471813867
77 0.89234865146429
78 0.893448315777289
79 0.894547980090288
80 0.892869545086237
81 0.893332561639079
82 0.892638036809816
83 0.893853455261026
84 0.893506192846394
85 0.893621946984605
86 0.895589767334182
87 0.895358259057761
88 0.894142840606552
89 0.895416136126867
90 0.893969209399236
91 0.894084963537446
92 0.893621946984605
93 0.894200717675657
94 0.895647644403287
95 0.89512675078134
96 0.89634216923255
97 0.89634216923255
98 0.89634216923255
99 0.89634216923255
100 0.895242504919551
};
\addlegendentry{val accuracy}
\addplot [semithick, forestgreen4416044]
table {%
1 0.415615233244588
2 0.416541266350272
3 0.635432341706216
4 0.652969093645098
5 0.646081722421577
6 0.642203958791527
7 0.644692672763051
8 0.646949878458155
9 0.647933788632944
10 0.645155689315893
11 0.651290658641046
12 0.650480379673573
13 0.653258478990624
14 0.64179881930779
15 0.64422965621021
16 0.642782729482579
17 0.642146081722422
18 0.638557703437898
19 0.646544738974418
20 0.640062507234634
21 0.625766871165644
22 0.63902071999074
23 0.631033684454219
24 0.627561060307906
25 0.635605972913532
26 0.628544970482695
27 0.638441949299687
28 0.627561060307906
29 0.634101169116796
30 0.633117258942007
31 0.66917467299456
32 0.679997684917236
33 0.67837712698229
34 0.670447968514874
35 0.675020256974187
36 0.675483273527029
37 0.676293552494502
38 0.677103831461975
39 0.679476791295289
40 0.676756569047343
41 0.680055561986341
42 0.683991202685496
43 0.67970829957171
44 0.675714781803449
45 0.677566848014816
46 0.672010649380715
47 0.672242157657136
48 0.68103947216113
49 0.684280588031022
50 0.675656904734344
51 0.674094223868503
52 0.684049079754601
53 0.682775784234286
54 0.685496006482232
55 0.67449936335224
56 0.673862715592082
57 0.682949415441602
58 0.682254890612339
59 0.680576455608288
60 0.679881930779025
61 0.684049079754601
62 0.68248639888876
63 0.675888413010765
64 0.678145618705869
65 0.682717907165181
66 0.676525060770923
67 0.680229193193657
68 0.661824285218197
69 0.653605741405255
70 0.666917467299456
71 0.665123278157194
72 0.66008797314504
73 0.670968862136821
74 0.66130339159625
75 0.662576687116564
76 0.664718138673458
77 0.666975344368561
78 0.6718370181734
79 0.671200370413242
80 0.660493112628777
81 0.660898252112513
82 0.640236138441949
83 0.672068526449821
84 0.659856464868619
85 0.6574256279662
86 0.666917467299456
87 0.664949646949878
88 0.669001041787244
89 0.641277925685843
90 0.634159046185901
91 0.642261835860632
92 0.658120152795462
93 0.654821159856465
94 0.674672994559556
95 0.671779141104294
96 0.615580507003125
97 0.615580507003125
98 0.615580507003125
99 0.615696261141336
100 0.605336265771501
};
\addlegendentry{test accuracy}
\end{axis}

\end{tikzpicture}

%% file: Images/1min_reliability_curve.tex
\begin{tikzpicture}

\definecolor{darkgray176}{RGB}{176,176,176}
\definecolor{darkorange25512714}{RGB}{255,127,14}
\definecolor{lightgray204}{RGB}{204,204,204}
\definecolor{steelblue31119180}{RGB}{31,119,180}

\begin{axis}[
clip=true,
height=0.8\linewidth,
legend cell align={left},
legend style={
  fill opacity=0.8,
  draw opacity=1,
  text opacity=1,
  at={(0.97,0.03)},
  anchor=south east,
  draw=lightgray204
},
scale only axis,
tick align=outside,
tick pos=left,
width=0.8\linewidth,
x grid style={darkgray176},
xlabel={predicted probability [\%]},
xmin=-0.05, xmax=1.05,
xtick style={color=black},
y grid style={darkgray176},
ylabel={observed frequency [\%]},
ymin=-0.05, ymax=1.05,
ytick style={color=black}
]
\addplot [line width=0.44pt, black, dotted, mark=*, mark size=1, mark options={solid}]
table {%
0 0
1 1
};
\addlegendentry{ideal}
\addplot [line width=0.44pt, steelblue31119180, mark=*, mark size=1, mark options={solid}]
table {%
3.70521941883494e-06 0
2.73363326890061e-05 0
9.37169109778445e-05 0
0.000350193014918348 0.00462962962962963
0.00124211469340026 0.0115740740740741
0.00445991678834003 0.0347222222222222
0.0172483427320822 0.0660486674391657
0.0608501012868967 0.159722222222222
0.188431332600165 0.292824074074074
0.431774925467072 0.480324074074074
0.689256741361556 0.628472222222222
0.858862924343517 0.733796296296296
0.937800072878217 0.841435185185185
0.970792236899549 0.88991888760139
0.985902594327384 0.935185185185185
0.993479160263917 0.953703703703704
0.997009333308598 0.976851851851852
0.998830056137138 0.989583333333333
0.999660481492047 1
0.999954461677621 1
};
\addlegendentry{uncalib.}
\addplot [line width=0.44pt, darkorange25512714, mark=*, mark size=1, mark options={solid}]
table {%
0.000334611942490971 0
0.00131628292353865 0
0.00285509809314741 0
0.00652487318049139 0.00462962962962963
0.0143836022119206 0.0115740740740741
0.0316026737808643 0.0347222222222222
0.0714169877645407 0.0660486674391657
0.149201243910953 0.159722222222222
0.282031462561485 0.292824074074074
0.456126730128498 0.480324074074074
0.624355937889103 0.628472222222222
0.758946741616411 0.733796296296296
0.847668033988115 0.841435185185185
0.901367655708891 0.88991888760139
0.935865456841717 0.935185185185185
0.959713644903596 0.953703703703704
0.97503783652732 0.976851851851852
0.986085411964359 0.989583333333333
0.993654101059759 1
0.998281148148357 1
};
\addlegendentry{calib.}
\end{axis}

\end{tikzpicture}

%% file: Images/1min_confidence_distribution.tex
\begin{tikzpicture}

\definecolor{darkgray176}{RGB}{176,176,176}
\definecolor{lightblue}{RGB}{173,216,230}
\definecolor{lightgray204}{RGB}{204,204,204}
\definecolor{steelblue31119180}{RGB}{31,119,180}

\begin{axis}[
clip=true,
height=0.8\linewidth,
legend cell align={left},
legend style={
  fill opacity=0.8,
  draw opacity=1,
  text opacity=1,
  at={(0.02,0.5)},
  anchor=west,
  draw=lightgray204
},
scale only axis,
tick align=outside,
tick pos=left,
width=0.8\linewidth,
x grid style={darkgray176},
xlabel={density [\%] (before ← | → after)},
xmin=-0.4, xmax=0.4,
xtick style={color=black},
y grid style={darkgray176},
ylabel={predicted probability [\%]},
ymin=0, ymax=1,
ytick style={color=black}
]
\draw[draw=none,fill=lightblue,fill opacity=0.7] (axis cs:0,0.0025) rectangle (axis cs:-0.369082069683991,0.0475);
\addlegendimage{ybar,ybar legend,draw=none,fill=lightblue,fill opacity=0.7}
\addlegendentry{uncalib.}

\draw[draw=none,fill=lightblue,fill opacity=0.7] (axis cs:0,0.0525) rectangle (axis cs:-0.0295751823127677,0.0975);
\draw[draw=none,fill=lightblue,fill opacity=0.7] (axis cs:0,0.1025) rectangle (axis cs:-0.0167843500405139,0.1475);
\draw[draw=none,fill=lightblue,fill opacity=0.7] (axis cs:0,0.1525) rectangle (axis cs:-0.0135432341706216,0.1975);
\draw[draw=none,fill=lightblue,fill opacity=0.7] (axis cs:0,0.2025) rectangle (axis cs:-0.0113439055446232,0.2475);
\draw[draw=none,fill=lightblue,fill opacity=0.7] (axis cs:0,0.2525) rectangle (axis cs:-0.0114596596828337,0.2975);
\draw[draw=none,fill=lightblue,fill opacity=0.7] (axis cs:0,0.3025) rectangle (axis cs:-0.0096075934714666,0.3475);
\draw[draw=none,fill=lightblue,fill opacity=0.7] (axis cs:0,0.3525) rectangle (axis cs:-0.00810278967473087,0.3975);
\draw[draw=none,fill=lightblue,fill opacity=0.7] (axis cs:0,0.4025) rectangle (axis cs:-0.00821854381294131,0.4475);
\draw[draw=none,fill=lightblue,fill opacity=0.7] (axis cs:0,0.4525) rectangle (axis cs:-0.00885519157309874,0.4975);
\draw[draw=none,fill=lightblue,fill opacity=0.7] (axis cs:0,0.5025) rectangle (axis cs:-0.00989697881699271,0.5475);
\draw[draw=none,fill=lightblue,fill opacity=0.7] (axis cs:0,0.5525) rectangle (axis cs:-0.00937608519504572,0.5975);
\draw[draw=none,fill=lightblue,fill opacity=0.7] (axis cs:0,0.6025) rectangle (axis cs:-0.00989697881699271,0.6475);
\draw[draw=none,fill=lightblue,fill opacity=0.7] (axis cs:0,0.6525) rectangle (axis cs:-0.0105915036462554,0.6975);
\draw[draw=none,fill=lightblue,fill opacity=0.7] (axis cs:0,0.7025) rectangle (axis cs:-0.0120384303738859,0.7475);
\draw[draw=none,fill=lightblue,fill opacity=0.7] (axis cs:0,0.7525) rectangle (axis cs:-0.014179881930779,0.7975);
\draw[draw=none,fill=lightblue,fill opacity=0.7] (axis cs:0,0.8025) rectangle (axis cs:-0.0169001041787244,0.8475);
\draw[draw=none,fill=lightblue,fill opacity=0.7] (axis cs:0,0.8525) rectangle (axis cs:-0.0243662460932978,0.8975);
\draw[draw=none,fill=lightblue,fill opacity=0.7] (axis cs:0,0.9025) rectangle (axis cs:-0.0428290311378632,0.9475);
\draw[draw=none,fill=lightblue,fill opacity=0.7] (axis cs:0,0.9525) rectangle (axis cs:-0.363352239842574,0.9975);
\draw[draw=none,fill=red,fill opacity=0.7] (axis cs:0,0.0025) rectangle (axis cs:0.303796735733302,0.0475);
\addlegendimage{ybar,ybar legend,draw=none,fill=red,fill opacity=0.7}
\addlegendentry{calib.}

\draw[draw=none,fill=red,fill opacity=0.7] (axis cs:0,0.0525) rectangle (axis cs:0.0442759578654937,0.0975);
\draw[draw=none,fill=red,fill opacity=0.7] (axis cs:0,0.1025) rectangle (axis cs:0.0288227804143998,0.1475);
\draw[draw=none,fill=red,fill opacity=0.7] (axis cs:0,0.1525) rectangle (axis cs:0.0217617779835629,0.1975);
\draw[draw=none,fill=red,fill opacity=0.7] (axis cs:0,0.2025) rectangle (axis cs:0.0165528417640931,0.2475);
\draw[draw=none,fill=red,fill opacity=0.7] (axis cs:0,0.2525) rectangle (axis cs:0.0158583169348304,0.2975);
\draw[draw=none,fill=red,fill opacity=0.7] (axis cs:0,0.3025) rectangle (axis cs:0.0144113902071999,0.3475);
\draw[draw=none,fill=red,fill opacity=0.7] (axis cs:0,0.3525) rectangle (axis cs:0.0148744067600417,0.3975);
\draw[draw=none,fill=red,fill opacity=0.7] (axis cs:0,0.4025) rectangle (axis cs:0.0122120615812015,0.4475);
\draw[draw=none,fill=red,fill opacity=0.7] (axis cs:0,0.4525) rectangle (axis cs:0.0140062507234634,0.4975);
\draw[draw=none,fill=red,fill opacity=0.7] (axis cs:0,0.5025) rectangle (axis cs:0.0150480379673573,0.5475);
\draw[draw=none,fill=red,fill opacity=0.7] (axis cs:0,0.5525) rectangle (axis cs:0.0151637921055678,0.5975);
\draw[draw=none,fill=red,fill opacity=0.7] (axis cs:0,0.6025) rectangle (axis cs:0.0163213334876722,0.6475);
\draw[draw=none,fill=red,fill opacity=0.7] (axis cs:0,0.6525) rectangle (axis cs:0.0170158583169348,0.6975);
\draw[draw=none,fill=red,fill opacity=0.7] (axis cs:0,0.7025) rectangle (axis cs:0.0200833429795115,0.7475);
\draw[draw=none,fill=red,fill opacity=0.7] (axis cs:0,0.7525) rectangle (axis cs:0.0239032295404561,0.7975);
\draw[draw=none,fill=red,fill opacity=0.7] (axis cs:0,0.8025) rectangle (axis cs:0.0307906007639773,0.8475);
\draw[draw=none,fill=red,fill opacity=0.7] (axis cs:0,0.8525) rectangle (axis cs:0.0480958444264382,0.8975);
\draw[draw=none,fill=red,fill opacity=0.7] (axis cs:0,0.9025) rectangle (axis cs:0.0775552726009955,0.9475);
\draw[draw=none,fill=red,fill opacity=0.7] (axis cs:0,0.9525) rectangle (axis cs:0.2494501678435,0.9975);
\addplot [steelblue31119180, forget plot]
table {%
0 0
0 1
};
\end{axis}

\end{tikzpicture}